\documentclass[sigconf, review=false]{acmart}
\renewcommand\footnotetextcopyrightpermission[1]{} 
\usepackage{booktabs,algorithm,algcompatible} 

\usepackage[english]{babel}
\usepackage{etoolbox}
\makeatletter
\patchcmd{\maketitle}{\@copyrightspace}{}{}{}
\makeatother

\usepackage[font=small,skip=0pt]{caption}
\setcopyright{none}

\makeatletter
\def\@copyrightspace{\relax}
\makeatother

\begin{document}
\title{A Self-Replication Basis For Designing Complex Agents}


\author{Thommen George Karimpanal}
\affiliation{%
  \institution{Singapore University of Technology and Design}
  \streetaddress{8 Somapah Road}
  \city{Singapore} 
  \postcode{487372}
}
\email{thommen_george@mymail.sutd.edu.sg}
\renewcommand{\shortauthors}{Karimpanal}

\begin{abstract}
In this work, we describe a self-replication-based mechanism for designing agents of increasing complexity. We demonstrate the validity of this approach by solving simple, standard evolutionary computation problems in simulation. In the context of these simulation results, we describe the fundamental differences of this approach when compared to traditional approaches. Further, we highlight the possible advantages of applying this approach to the problem of designing complex artificial agents, along with the potential drawbacks and issues to be addressed in the future. 
\end{abstract}
%
%


\copyrightyear{2018}
\acmYear{2018}
\setcopyright{none}
\acmConference[GECCO '18 Companion]{Genetic and Evolutionary
Computation Conference Companion}{July 15--19, 2018}{Kyoto, Japan}
\acmBooktitle{GECCO '18 Companion: Genetic and Evolutionary
Computation Conference Companion, July 15--19, 2018, Kyoto,
Japan}\acmDOI{10.1145/3205651.3208762}
\acmISBN{978-1-4503-5764-7/18/07}

\maketitle

\section{Introduction}

Since the birth of artificial intelligence, the ultimate aim of the field has been to replicate intelligent behavior akin to those found in humans and animals. To this end, a number of tools and techniques have been developed. Among these, reinforcement learning \cite{sutton1998reinforcement}, deep learning \cite{lecun2015deep} and evolutionary algorithms \cite{koza1994genetic,holland1992adaptation} are the most powerful and popular ones till date. Although these methods are extremely useful, they require some form of explicit task specification. As a result, they have a limited degree of autonomy, and fail to generalize well \cite{narodytska2016simple,taylor2009transfer}. In addition, the mentioned approaches fail to capture the mechanisms through which intelligent beings came into existence in natural systems. Even approaches that are inspired from biological evolution require the specification of a fitness function, and the aim is solely to solve a specific optimization problem. Even when the aim is to simulate artificial life \cite{sims1994evolving,nolfi2016evolutionary}, these approaches do not capture the trends towards increasing diversity and complexity which are seen in biological evolution.  
We hypothesize that in order to design truly intelligent agents, the fundamental, natural processes that led to their creation need to be recreated to a certain extent. In this work, we propose an approach based on the zero-force evolutionary law \cite{mcshea2010biology}, which states that an increase in diversity and complexity is a natural outcome of evolutionary systems that possess the properties of heredity and variation. 
These properties constitute an imperfect self-replication process, which is the key idea behind our approach. 

\section{Approach and Preliminary Results}
\label{methods}
Our approach starts with a population of fundamental elements, analogous to a primordial soup \cite{haldane1929rationalist} which contains  the fundamental components needed to build more complex entities/agents. However, in each generation, such an agent is allowed to survive and self-replicate only when a certain replication rule is followed. This rule is similar to the fitness function used in traditional evolutionary algorithms, in the sense that it determines which agents are allowed to continue to the next generation. However, unlike traditional approaches, the selection is not fitness proportional. Instead, agents that follow the replication rule are allowed to replicate and those that do not are removed from the population. Each agent is assigned a limited number of generations, referred to as the generation lifetime of an agent, within which it may self-replicate. Once the generation lifetime decays to zero, it is removed from the population. The imperfect nature of self-replication allows mutations to occur with a fixed, pre-defined probability, otherwise producing identical offspring. The nature of the mutation can be additive or subtractive, and occur with equal probability. Additive mutations append the offspring with new, randomly picked elements, while subtractive mutations remove a randomly picked element from the offspring agents. This allows the number of elements comprising the agent (agent complexity) to grow or reduce across generations.
The proposed algorithm is summarized in Algorithm \ref{alg:algorithm1}.

\begin{algorithm}[h!]
  \caption{Emergence of complex agents using self-replication}
  \begin{algorithmic}[1]
    \STATE \textbf{Inputs}: \STATEx $G_{max}:$~Maximum number of generations \STATEx $L:$~ Maximum generation lifetime
    \STATEx $l_{i}$~Generation lifetime of agent $i$
    \STATEx $N_{a}:$~Population of agents
    \STATEx
    $P_{m}:$~Probability of mutation \STATEx $E:$~Set of fundamental elements
    \STATE Initialize population with $N_{a}$ agents using elements from $E$
    \FOR {$i=1:G_{max}$}
    \FOR {$j=1:N_{a}$}
    \IF {agent $j$ satisfies the replication rule}
    \STATE Self-replicate (with $P_{m}$ probability of mutated offspring)
    \STATE Set initial generation lifetime of offspring to be $L$
    \STATE $l_{j}=l_{j} -1$
    \ELSE {Remove agent $j$ from population}
    \ENDIF
    \ENDFOR 
    \STATE Remove agents with generation lifetime $l \leq 0$
    \STATE Update $N_{a}$
    \ENDFOR
  \end{algorithmic}
  \label{alg:algorithm1}
\end{algorithm}
In order to demonstrate the described approach, we consider a problem of discovering the sequence of all prime numbers up to a given number $N$. The set of fundamental elements $E$ is thus the set of integers from $1$ to $N$. Here, the hyperparameters mentioned in Algorithm \ref{alg:algorithm1} are set to be as follows: $N=100$, $G_{max}=500$, $P_{m}=0.2$, $N_{a}=100$ and $L=4$. The $100$ agents are initialized to be a random integer between $1$ and $100$, and are allowed to self-replicate as described in Algorithm \ref{alg:algorithm1}. The rule for self replication in this case is simply that the agents must be a continuous sequence of prime numbers starting from $2$, without repetition. With this replication rule, initially, agents of complexity $1$ (here, the length of the sequence is synonymous with complexity) are discovered, and they replicate, leading to an exponential growth in population. Subsequently, owing to mutations, more complex agents are discovered, and are allowed to replicate. This process continues until the sequence of all prime numbers $\leq N$ is discovered. In the final population, agents with lower complexities emphatically outnumber those with higher complexities. This is a feature that is also true in biological ecosystems, perhaps due to the similar manner in which complex species evolve from simpler ones.
In practice, since the growth of the agents is so rapid, and since the algorithm loops through all the agents in the population, the discovery of more complex agents eventually becomes prohibitively slow and computationally intensive. In order to overcome this limitation, one may periodically eliminate agents with lower levels of complexity, and focus the computational effort on more complex agents. With this periodic selective extinction approach, using an ordinary desktop computer, the complete sequence of primes numbers was obtained in the order of a minute's time.
We also applied this approach to the \emph{OneMax} problem, 
in which the objective is to maximize the number of $1$'s in a fixed length string of numbers. To this end, the replication rule of the prime number problem was merely modified to the following: if the agent has all elements as $1$, allow it to replicate; if not, remove it from the population.
\begin{figure}[ht]
\centering
\includegraphics[width=1\linewidth]{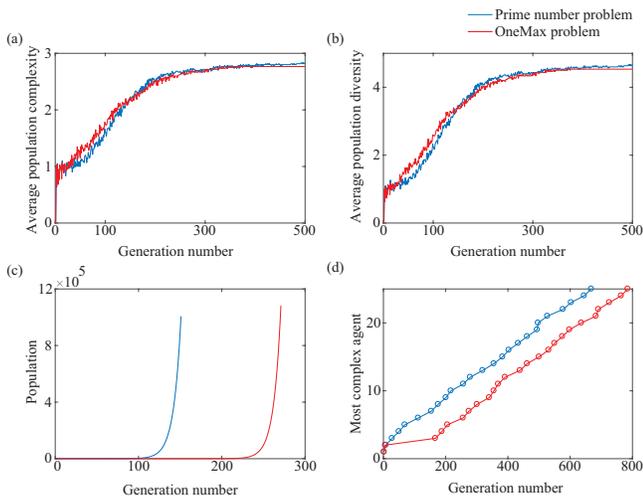}
\caption{(a) and (b) show the average increases in complexity and diversity of the population over $30$ runs, with the number of generations. (c) shows the typical trend of the population when no extinction event is enforced. (d) shows the typical trend of the maximum complexity of a population when periodic (whenever total population exceeded $10^{6}$ agents) extinction events are enforced.
} 
\label{fig:results}
\end{figure} 

Figures \ref{fig:results}(a) and \ref{fig:results}(b) show that the described approach leads to increased complexity and diversity with the number of generations. The exponential increase in the population (Figure \ref{fig:results}(c)) makes the computation intractable in the absence of extinction events. As a result, only a complexity of up to $5$ could be achieved as shown in Figures \ref{fig:results}(a) and \ref{fig:results}(b). However, periodic forced extinction events allow more complex solutions to be discovered at a steady rate, as shown in Figure \ref{fig:results}(d). This shows that the proposed approach, apart from being able to evolve agents of increasing complexity, can also be used as a stochastic optimization tool.
 
In nature, replicative success is determined by specific conditions imposed by the environment itself. Hence, in general, designing appropriate replication rules may not be trivial, as in the case of fitness functions. However, the less restrictive nature of the replication rule may allow for greater flexibility when compared to traditional evolutionary approaches. 
Although our approach, as described, does not include a learning component, this can be incorporated into the innermost `for' loop in Algorithm \ref{alg:algorithm1}. This opens up the possibility of incorporating established learning approaches and leveraging the Baldwin effect \cite{baldwin1896new} to guide the evolutionary process. Doing so could possibly constitute a more realistic approach for designing truly autonomous and intelligent agents.

\section{Conclusions}
In this work, we introduced an approach that captures the typical trends of increasing complexity and diversity observed in biological evolution. We tested the approach on two simple problems, and in this context, described its important characteristics. We also proposed a method to simply utilize this approach as a tool for solving stochastic optimization problems. We posit that such an approach, especially when combined with existing learning techniques, could enable the design of truly intelligent artificial agents. 

\bibliographystyle{ACM-Reference-Format}
\bibliography{sample-bibliography} 

\end{document}